\documentclass[nohyperref]{article}

\usepackage{microtype}
\usepackage{graphicx}
\usepackage{subfigure}
\usepackage{booktabs} 

\usepackage{hyperref}

\usepackage[accepted]{icml2023}

\usepackage{amsmath}
\usepackage{amssymb}
\usepackage{mathtools}
\usepackage{amsthm}

\usepackage[capitalize,noabbrev]{cleveref}

\usepackage{multirow}
\usepackage{rotating,tabularx}

\theoremstyle{plain}

\theoremstyle{definition}

\theoremstyle{remark}

\newcommand{\Fullname}{Class Typical Matching}

\newcommand{\shortname}{CTM}

\usepackage[disable,textsize=tiny]{todonotes}

\icmltitlerunning{A Cosine Similarity-based Method for Out-of-Distribution Detection}

\begin{document}

\twocolumn[
\icmltitle{A Cosine Similarity-based Method for Out-of-Distribution Detection}



\icmlsetsymbol{equal}{*}

\begin{icmlauthorlist}
\icmlauthor{Nguyen Ngoc-Hieu}{fpt}
\icmlauthor{Nguyen Hung-Quang}{vinuni}
\icmlauthor{The-Anh Ta}{fpt}
\icmlauthor{Thanh Nguyen-Tang}{jhu}
\icmlauthor{Khoa D. Doan}{vinuni}
\icmlauthor{Hoang Thanh-Tung}{fpt}
\end{icmlauthorlist}

\icmlaffiliation{fpt}{FPT Software AI Center}
\icmlaffiliation{vinuni}{VinUniversity}
\icmlaffiliation{jhu}{Johns Hopkins University}

\icmlcorrespondingauthor{Nguyen Ngoc-Hieu}{ngochieutb13@gmail.com}

\icmlkeywords{Machine Learning, ICML}

\vskip 0.3in
]



\printAffiliationsAndNotice{}  

\begin{abstract}
    The ability to detect OOD data is a crucial aspect of practical machine learning applications.
    In this work, we show that cosine similarity between the test feature and the typical ID feature is a good indicator of OOD data.
    We propose \Fullname\;(\shortname), a post hoc OOD detection algorithm that uses a cosine similarity scoring function.
    Extensive experiments on multiple benchmarks show that \shortname\;outperforms existing post hoc OOD detection methods.
\end{abstract}

\section{Introduction}
In machine learning, distribution shift is the problem where the test distribution is not identical to the training distribution.
Deep learning (DL) models, including those with good i.i.d. generalization performance, often perform poorly when distribution shifts occur \cite{Nguyen2014}.
In real-world applications, distribution shifts are unavoidable because the environment changes in time.
An emerging requirement for DL systems is that they must be able to handle distribution shifts \cite{AmodeiOSCSM16}. 

A common approach to the distribution shift problem is to detect out-of-distribution (OOD) samples, samples from the shifted distribution, and remove them from the test data.
There is a wide array of OOD detection methods, ranging from classification-based, and density-based to distance-based methods \cite{Yang2021Survey}. 
Classification-based methods, which classify incoming data as OOD or in-distribution (ID) based on the confidence or feature embedding assigned by a classifier, are some of the most commonly used methods. 
Several improvements to classification methods have been proposed, including modifying the loss function \cite{wei2022logitnorm, ming2023how}, changing the classifier architecture \cite{Malinin2018Predictive}, and using post hoc processing techniques \cite{hendrycks17baseline, Liang2017, liu2020energy, mahalanobis}. 
Among these techniques, post hoc methods are often preferred in practice due to their simplicity and ease of integration with pre-trained models without the need for additional training. 

In this paper, we introduce \Fullname\;(\shortname), a post hoc algorithm for OOD detection.
\shortname\; is based on our observation that the cosine similarity between the test input's feature and the in-distribution features is very useful for OOD detection (Section \ref{sec:Motivation}).
Different from other post hoc methods such as \texttt{Mahalanobis} \cite{lee2018mahaood} and \texttt{KNN} \cite{sun2022knn} which leverage Euclidean distance in the feature space, our method uses cosine similarity for OOD score computation.
In section \ref{sec:analysis}, we theoretically show that cosine similarity is a good indicator for OOD samples.
Our contributions are as follows:
\begin{enumerate}
    \item We empirically and theoretically show that cosine similarity between the feature representation of a test input and a typical ID feature is an effective scoring function for OOD detection.
    \item We propose \shortname, a post hoc method that uses angular information for improved OOD detection.
    \item We perform extensive experiments and ablation studies to evaluate the proposed method across 3 ID datasets and 10 OOD datasets.
\end{enumerate}

\section{Method}
\label{sec:method}
\subsection{Prelimaries}
\paragraph{Problem statement.}
We denote $\mathcal{X} \subseteq \mathbb{R}^{d}$ the input space and $\mathcal{Y} = \{y_1, \dots, y_C\}$ the label space. 
A classifier $f: \mathcal{X} \mapsto \mathbb{R}^{C}$ learns to map a given input $\mathbf{x} \in \mathcal{X}$ to the output space.
Let $p_{\text{train}}(\mathbf{x}, y)$ denote a probability distribution defined on $\mathcal{X} \times \mathcal{Y}$.
Further more, let $p_{\text{train}}(\mathbf{x})$ and $p_{\text{train}}(y)$ denote the marginal probability distribution on $\mathcal{X}$ and $\mathcal{Y}$, respectively. 
The goal is to design a binary function estimator $g: \mathcal{X} \to \{\text{in}, \text{out}\}$ that classifies whether a test example $\mathbf{x}$ is generated from $p_{\text{train}}(\mathbf{x})$ or not.

The concept of distribution shift is very diverse. It presents a challenge because being robust against one type of shift does not mean it will be effective against another shift. 
Therefore, it is important to characterize real-world shifts in order to develop effective methods for mitigating their impact.
In this paper, we work with the image data and adopt an experimental setting presented in previous works where OOD samples are drawn from unknown classes \cite{fang2022is, Yang2021Survey}.

One common approach to out-of-distribution (OOD) detection is to construct a scoring function $S: \mathcal{X} \mapsto \mathbb{R}$ that assigns lower scores to points drawn from an out-distribution  $q(\mathbf{x})$. The detector, denoted as $g$, is then constructed based on the level set obtained from the score function
\begin{equation*}
g(\mathbf{x})= \begin{cases}\text { ID}, & \text { if } S(\mathbf{x}) \ge \lambda \\ \text { OOD}, & \text { if } S(\mathbf{x}) < \lambda\end{cases},
\end{equation*}
where $S(\mathbf{x})$ denotes a scoring function and $\lambda$ is commonly set so that $g$ correctly classifies a high proportion (e.g., 95\%) of in-distribution (ID) data.

\paragraph{Notation.}
Posthoc methods often use a trained neural network to derive the score function.
A trained deep NN classifier generally consists of two components: 
(1) a deep feature extractor that maps the input to a feature embedding and 
(2) a head that maps the embedding to an output.
The most common choice for feature embedding is the output of the penultimate layer just before the classification layer. 
We denote the feature embedding map by $h: \mathcal{X} \mapsto \mathbb{R}^{m}$, where 
$m$ is the dimension of the embedding. 
Given $\mathbf{x} \in \mathcal{X}$, denote $\mathbf{z} \in \mathbb{R}^{m}$ the feature of $\mathbf{x}$ i.e. $\mathbf{z} = h(\mathbf{x})$.
The last FC layer in a neural network is given by:
\begin{equation}
    f(\mathbf{x}) = W h(\mathbf{x}) + \mathbf{b} = W \mathbf{z} + \mathbf{b}
\end{equation}
where $W \in \mathbb{R}^{C \times m}$ is the weight matrix, and $\mathbf{b} \in \mathbb{R}^{C}$ is the bias vector.
We also denote $\mathbf{w}_k$ the $k$-th row of $W$.
For operators, we denote $\left<\cdot, \cdot \right>$ the inner product between two vectors and $\|\cdot \|$ is the Euclidean norm. 

\subsection{Method}
\label{sec:Motivation}
\citet{hendrycks2022scaling} show that the Maximum logit score is a strong baseline for large-scale OOD detection.
This method scores each input by the largest values of their logits vector. 
Concretely, given the input's penultimate feature $\mathbf{z}$ the score is computed by the following equation: 
\begin{equation*}
    \underset{k}{\max} \left<\mathbf{w}_k, \mathbf{z}\right> + b_k
\end{equation*}
where $\mathbf{w}_k$ and $b_k$ are weights and bias of the last layer w.r.t class $k$. 
This score function can also be formulated as 
\begin{equation*}
    \max_k \|\mathbf{w}_k\|\|\mathbf{z}\|\cos (\mathbf{w}_k, \mathbf{z}) + b_k
\end{equation*}
where $\cos (\mathbf{w}_k, \mathbf{z}) $ is the cosine similarity between $\mathbf{w}_k$ and $\mathbf{z}$.
This formulation separates norm terms ($\|\mathbf{w}_k\|$, $\|\mathbf{z}\|$) and angular term $\cos (\mathbf{w}_k, \mathbf{z}) $. 
Note that for a particular input, the model's prediction is ${\arg\max}_k \;\|\mathbf{w}_k\|\|\mathbf{z}\|\cos (\mathbf{w}_k, \mathbf{z}) + b_k. $
As $\|\mathbf{z}\|$ is fixed for different $k$ and the terms $\|\mathbf{w}_k\|$ and $b_k$ are independent of the input, the cosine similarity term $\cos (\mathbf{w}_k, \mathbf{z})$ carries the most information for the model's prediction. When the $\cos (\mathbf{w}_k, \mathbf{z}) $ values are similar for different $k$ values, the score is influenced by the norms of $\mathbf{w}_k$ and $b_k$. 
This can make the OOD problem more challenging if the OOD sample is assigned to a class with larger weight norm $\|\mathbf{w}_k\|$ and bias $b_k$.
We also observe that the norm of OOD feature embeddings can be large and increase the score.
In fact, \citet{sun2022knn} found that using the normalized penultimate feature greatly improves the \texttt{KNN} method, while
\citet{wei2022logitnorm} suggests that the norm of \textit{logit} is the source of the over-confident behavior of neural network trained with cross-entropy loss. 
\texttt{CIDER} \cite{ming2023how} uses hypersphere representation to benefit OOD detection tasks by designing an end-to-end loss function.
We argue that using only the term $\cos (\mathbf{w}_k, \mathbf{z})$ can retain the performance on the OOD detection task. 
Furthermore, our empirical finding suggests that replacing $\mathbf{w}_k$ with $\boldsymbol{\mu}_k$ - the mean of feature vectors in class $k$, and using $\cos(\boldsymbol{\mu}_k, \mathbf{z})$ improve OOD detection performance. Intuitively, this score can be thought of as computing the similarity between the input's feature and the typical feature of class $k$.  

From the above motivation, we propose using cosine similarity with within-class feature mean $\boldsymbol{\mu}_k$ for OOD detection
\begin{equation*}
g(\mathbf{x})= \begin{cases}\text { ID}, & \text { if } \underset{k}{\max} \cos(\boldsymbol{\mu}_k, \mathbf{z}) \ge \lambda \\ \text { OOD}, & \text { otherwise} \end{cases},
\end{equation*}
where $\lambda$ is the threshold.
The score function $S(\mathbf{x}) = \max_k \cos(\boldsymbol{\mu}_k, \mathbf{z})$ measures the similarity between the test input's feature and within-class mean features. 
In the next section, we show that this simple idea is, in fact, very effective for detecting OOD inputs.
\begin{table*}[ht!]
\centering
\caption{\textbf{OOD detection results on ImageNet.} Proposed and baseline methods are based on a ResNet-50 \cite{he2016deep} model trained on ImageNet-1k \cite{deng2009imagenet} only. $\downarrow$ indicates smaller values are better and $\uparrow$ indicates larger values are better.}
\label{tab:imagenet_result}
\resizebox{1.0\textwidth}{!}
{
    \begin{tabular}{c c c c c c c c c c c}
    \hline
     & \multicolumn{8}{c}{\textbf{OOD Datasets}} & & \\
    \cline{2-9} \\
    \textbf{Methods} & \multicolumn{2}{c}{\textbf{iNaturalist}} & \multicolumn{2}{c}{\textbf{SUN}} & \multicolumn{2}{c}{\textbf{Places}} & \multicolumn{2}{c}{\textbf{Textures}} & \multicolumn{2}{c}{\textbf{Average}} \\
    & & & & & & & & & & \\
    & FPR95 & AUROC & FPR95 & AUROC & FPR95 & AUROC & FPR95 & AUROC & FPR95 & AUROC \\
    & $\downarrow$ & $\uparrow$ & $\downarrow$ & $\uparrow$ & $\downarrow$ & $\uparrow$ & $\downarrow$ & $\uparrow$ & $\downarrow$ & $\uparrow$ \\
    \midrule
    \multicolumn{1}{l}{Softmax score~\cite{hendrycks17baseline}} & 54.99 & 87.74 & 70.83 & 80.86 & 73.99 & 79.76 & 68.00 & 79.61 & 66.95 & 81.99 \\
    \multicolumn{1}{l}{MaxLogit~\cite{hendrycks2022scaling}} & 50.78 & 91.15 & 60.42 & 86.44 & 66.07 & 84.03 & 54.93 & 86.39 & 58.05 & 87.00 \\
    \multicolumn{1}{l}{ODIN~\cite{Liang2017}} & 47.66 & 89.66 & 60.15 & 84.59 & 67.89 & 81.78 & 50.23 & 85.62 & 56.48 & 85.41 \\
    \multicolumn{1}{l}{Mahalanobis~\cite{mahalanobis}} & 97.00 & 52.65 & 98.50 & 42.41 & 98.40 & 41.79 & 55.80 & 85.01 & 87.43 & 55.47 \\
    \multicolumn{1}{l}{Energy score~\cite{liu2020energy}} & 55.72 & 89.95 & 59.26 & \textbf{85.89} & 64.92 & \textbf{82.86} & 53.72 & 85.99 & 58.41 & 86.17 \\
    \multicolumn{1}{l}{KNN~\cite{sun2022knn}} & 59.08 & 86.20 & 69.53 & 80.10 & 77.09 & 74.87 & \textbf{11.56} & \textbf{97.18} & 54.32 & 84.59\\
    \cmidrule{1-11}
    \multicolumn{1}{l}{\textbf{\shortname\;(Our)}} & \textbf{22.58} & \textbf{95.51} & \textbf{55.02} & 85.55 & \textbf{63.07} & 81.73 & 15.25 & 96.70 & \textbf{38.98} & \textbf{89.87}\\
    \bottomrule
    \end{tabular}
}
\vspace{0.1in}

\end{table*}
\section{Experiments}

In this section, we present the experimental results of \shortname\; on several benchmarks
and an ablation study of the method.

\subsection{Experimental Setup}
\label{sec:exp_set}
\paragraph{Datasets and models.} We conducted experiments on moderate and large-scale benchmarks. 
The moderate benchmarks include CIFAR-10 and CIFAR-100 \cite{krizhevsky2009learning} as in-distribution datasets and six out-of-distribution datasets: SVHN \cite{netzer2011reading}, LSUN-Crop \cite{yu2015lsun}, LSUN-Resize \cite{yu2015lsun}, iSUN \cite{xu2015turkergaze}, Textures \cite{cimpoi2014describing}, and  Places365 \cite{zhou2017places}.
The model used in the CIFAR benchmarks is a pre-trained DenseNet-101 \cite{huang2017densely}.
The proposed method was also evaluated on a large-scale dataset, using ImageNet-1k \cite{deng2009imagenet} as the ID dataset and four OOD datasets: iNaturalist \cite{van2018inaturalist}, SUN \cite{xiao2010sun}, Place365 \cite{zhou2017places}, and Textures \cite{cimpoi2014describing}.
We used ResNet-50 \cite{he2016deep} as the backbone for the ImangeNet benchmark. 
All networks were pre-trained using ID datasets without regularizing on auxiliary outlier data. 
The model parameters remained unchanged during the OOD detection phase, providing a fair comparison among the different methods. 

\paragraph{Evaluation metrics.}
In our study, we evaluated the performance of OOD detection by measuring the following metrics: (1) the False Positive Rate (FPR95), which is the percentage of OOD images that were wrongly classified as ID images when the true positive rate of ID examples is 95\%; (2) the Area Under the Receiver Operating Characteristic curve (AUROC), which assesses the overall performance of the OOD detection method; and (3) the Area Under the Precision-Recall curve (AUPR).

\subsection{Results on both CIFAR and ImageNet benchmarks}
\label{sec:exp_results}
We compared our method with other established post-hoc methods that do not require modifying the training process and have similar computational complexity and time requirements. Specifically, we selected \texttt{MSP}~\cite{hendrycks17baseline}, \texttt{MaxLogit}~\cite{hendrycks2022scaling}, \texttt{Energy}~\cite{liu2020energy}, \texttt{ODIN}~\cite{Liang2017}, \texttt{Mahalanobis}~\cite{mahalanobis}, and \texttt{KNN}~\cite{sun2022knn}. The results of the CIFAR evaluations are presented in Table \ref{tab:main_cifar_results}. We report the average performance over the six OOD datasets for 2 evaluation metrics: FPR95 and AUROC. 
On average \shortname\; has 96.40\% AUROC on CIFAR-10 and 89.11\% AUROC on CIFAR-100, which is competitive with \texttt{KNN} method on CIFAR-10 and surpasses it on CIFAR-100 benchmark while algorithmically simpler.
Detailed performance on individual datasets is reported in Appendix \ref{appsec:detail_cifar}.

\begin{table}[hbt!]
\caption{\textbf{OOD detection results on CIFAR benchmarks.} The results were  averaged from 6 OOD datasets and measured in terms of FPR95 and AUROC. All values are percentages. All methods are based on a DenseNet-101 \cite{huang2017densely} model trained on ID data only.}
\label{tab:main_cifar_results}
\centering 
\scalebox{0.8}{
\begin{tabular}{l c c | c c}
\toprule
 & \multicolumn{2}{c}{\textbf{CIFAR-10}} & \multicolumn{2}{c}{\textbf{CIFAR-100}} \\
 \cmidrule(r){2-5}
\textbf{Method} & \textbf{FPR95} {$\downarrow$} & \textbf{AUROC} $\uparrow$ & \textbf{FPR95} $\downarrow$ & \textbf{AUROC} $\uparrow$ \\
\midrule
Softmax score & 48.73 & 92.46 & 80.13 & 74.36 \\
MaxLogit & 26.44 & 94.47 & 69.98 & 80.31 \\
Energy score & 26.55 & 94.57 & 68.45 & 81.19 \\
ODIN & 24.57 & 93.71 & 58.14 & 84.49 \\
Mahalanobis & 31.42 & 89.15 & 55.37 & 82.73 \\
KNN & \textbf{16.61} & \textbf{ 96.71} & 42.34 & 87.56 \\
\midrule
\textbf{\shortname\;(Ours)} & 18.23 & 96.40 & \textbf{41.76} & \textbf{89.11} \\
\bottomrule
\end{tabular}
}
\vspace{0.1in}
\end{table}

Table \ref{tab:imagenet_result} presents the performance of OOD detection methods on ImageNet. 
As we can see, \shortname\;establishes favorable performance across OOD datasets and evaluation metrics. It reduces the FPR95 metrics by 15.43\% compare to \texttt{KNN}. 

\subsection{Cosine similarity is informative.}
\label{sec:ablation}
We test the effectiveness of cosine similarity for OOD detection by making two modifications to the prediction process of an already trained network, (1) remove the bias $b_k$ and normalize $\mathbf{w}_k$ and (2) normalize the penultimate features before feeding them to the linear layer. The prediction function after these modifications is given by the following equation:
\begin{equation*}
    \underset{k}{\arg\max} \frac{\exp{ \left<\hat{\mathbf{w}}_k, \hat{\mathbf{z}}\right> }}{\sum_{c} \exp{ \left<\hat{\mathbf{w}}_c , \hat{\mathbf{z}}\right> }} = \underset{k}{\arg\max} \frac{\exp{ \cos (\mathbf{w}_k, \mathbf{z}) }}{\sum_{c} \exp{ \cos\theta_c(\mathbf{z})  }}
\end{equation*}
where $\hat{\mathbf{w}}_k = \frac{\mathbf{w}_k}{\|\mathbf{w}_k\|}$ and $\hat{\mathbf{z}} = \frac{\mathbf{z}}{\|\mathbf{z}\|}$.
We call this modification \textbf{CW} stands for cosine with weight. 
CW discards the magnitude component in $\mathbf{z}$ and the prediction is solely based on its direction. 
We also present another modification: cosine with mean (\textbf{CM}) which replaces $\mathbf{w}_k$ by the mean $\boldsymbol{\mu}_k$ of the training feature of class $k$.
For this experiment we use two architecture: WideResNet-40-2 \cite{zagoruyko2016wide} and DenseNet-101 \cite{huang2017densely}, and two datasets: CIFAR-10 and CIFAR-100.  
The result, reported in table \ref{tab:acc_angular_pred}, suggests that using only angular information $\cos (\mathbf{w}_k, \mathbf{z})$ can retain the most performance of the OOD detection task without much degradation on the classification task. 
Notice that CM increases the OOD detection performance by a large margin. 
These results verify our motivation in section \ref{sec:Motivation} that cosine similarity is informative for both classifying ID examples and detecting OOD examples.

\begin{table}[ht]
  \caption{\textbf{Cosine similarity is effective.} Test accuracy and OOD Detection performance (AUROC) of models before and after modification.}
  \label{tab:acc_angular_pred}
  \centering
  \scalebox{0.72}{
  \begin{tabular}{ccc}
    \toprule
    \multirow{2}{*}{Model \& Dataset} & \textbf{Test Accuracy}  & \textbf{AUROC} \\ 
    \cmidrule(r){2-3} & \textbf{Standard/CW/CM} & \textbf{Standard/CW/CM} \\ 
    \midrule
    WideResNet-40-2 + CIFAR-10       & 94.84/94.82/\textbf{95.02}                              & 91.29/\textbf{92.49}/\textbf{92.49} \\ 
    WideResNet-40-2 + CIFAR-100      & \textbf{75.95}/75.93/75.03                              & 77.39/79.77/\textbf{86.95} \\ 
    DenseNet + CIFAR-10              & 94.52/\textbf{94.55}/94.40                              & 94.62/94.40/\textbf{96.40} \\ 
    DenseNet + CIFAR-100             & \textbf{75.08}/74.69/71.66                              & 80.28/75.01/\textbf{89.11} \\ 
    \bottomrule
  \end{tabular}
  }
\end{table}

\section{An analysis of cosine similarity from influence perspective}
\label{sec:analysis}
In this section, we analyze why using cosine similarity between the input feature and the mean is effective for out-of-distribution detection. 
We show that cosine similarity naturally arises from the influence perspective, which characterizes how a function's value at one input changes when we modify its value at another input.
In particular, given a scalar output function $g_W$ parameterized by $W$, \citet{charpiat2019input} proposes a kernel measuring the influence between two input $\mathbf{z}$ and $\mathbf{z}'$: 
\begin{equation*}
    K_{g}(\mathbf{z}, \mathbf{z}') = \frac{\left< \nabla_W g_W(\mathbf{z}),  \nabla_W g_W(\mathbf{z}') \right>}{\| \nabla_W g_W(\mathbf{z})\| \|  \nabla_W g_W(\mathbf{z}')\|}.
\end{equation*}
This kernel measures how similar output of $g$ at $\mathbf{z}$ and $\mathbf{z}'$ change if the weight $W$ is perturbed.
Notice that the kernel is bounded between $[-1, 1]$.
If the value $K_g(\mathbf{z}, \mathbf{z}')$ closes to $1$ then $g_W(\mathbf{z})$ and $g_W(\mathbf{z}')$ response similar to each other for a perturbation on $W$.
Intuitively, large $K_g(\mathbf{z}, \mathbf{z}')$ indicates that $\mathbf{z}$ and $\mathbf{z}'$ are similar under the point of view of $g$.

For the choice of $g$, inspired by \citet{huang2021on}, we let $g$ be the Kullback–Leibler (KL) divergence between the softmax output and a uniform distribution. 
Formally, denote $\mathbf{p} = \operatorname{softmax}(W\mathbf{z} + \mathbf{b})$ and $\mathbf{p}' = \operatorname{softmax}(W\mathbf{z}' + \mathbf{b})$ as predicted label probabilities assign to feature $\mathbf{z}$ and $\mathbf{z}'$ by the trained model.
Then we have $g = D_{\text{KL}} (\mathbf{u} || \mathbf{p})$, where $D_{\text{KL}}$ is the KL divergence,
and $\mathbf{u}$ is a uniform distribution on labels, i.e $\mathbf{u}= [1/C,1/C,\dots,1/C] \in \mathbb{R}^C$.
The gradient of $g(\mathbf{z}')$ w.r.t $W$ points to direction which increase model's uncertainty at $\mathbf{z}'$.
The kernel $K_g$ now measure how much the predicted distribution at $\mathbf{z}$ become uniform if the weight $W$ is perturbed such that increasing the uncertainty at $\mathbf{z}'$.
Intuitively, if $\mathbf{z}'$ is a typical ID point then this perturbation will affect the model's prediction on an ID input more than on an OOD input. This also makes $K_g$ agnostic to the true label of $\mathbf{z}$ or $\mathbf{z}'$.

Given the kernel $K_g$, for each test point $\mathbf{z}$ which is predicted with label $k$, we chose the point $\mathbf{z}'$ as the reference point such that it represents class $k$ and measure the influence between $\mathbf{z}'$ and $\mathbf{z}$. 
To get the value $K_g(\mathbf{z}, \mathbf{z}')$, we first compute the gradient of $g$ w.r.t.~$W$. 
We have
\begin{align*}
    \nabla_W g_W (\mathbf{z}) 
    &= \frac{\partial D_{\text{KL}} (\mathbf{u} || \mathbf{p}_{W})}{\partial \operatorname{vec}{W}} \\
    &= (\mathbf{u} - \mathbf{p})^\top  \otimes \mathbf{z} ^\top \in \mathbb{R}^{C d}.
\end{align*}
where $\otimes$ is Kronecker product. 
Assume that $\mathbf{z}$ is not a zero vector and $\mathbf{p}$ is not uniform then $\|\nabla_W g_W (\mathbf{z}) \| > 0$. Apply the same assumption for $\mathbf{z}'$ and $\mathbf{p}'$ then 
\begin{align}
    K_g (\mathbf{z}', \mathbf{z}) &= \frac{\left< \nabla_W g_W(\mathbf{z}'),  \nabla_W g_W(\mathbf{z}) \right>}{\| \nabla_W g_W(\mathbf{z}')\| \|  \nabla_W g_W(\mathbf{z})\|} \nonumber \\
    &= \frac{\left< \mathbf{u} - \mathbf{p}', \mathbf{u} - \mathbf{p} \right>}{\| \mathbf{u} - \mathbf{p}' \|\| \mathbf{u} - \mathbf{p}\|} \cdot \frac{\left< \mathbf{z}', \mathbf{z} \right>}{\|\mathbf{z}'\|\|\mathbf{z}\|} \nonumber \\ 
    &= \frac{\left<\mathbf{p}',\mathbf{p} \right> - 1/C}{\| \mathbf{u} - \mathbf{p}' \|\| \mathbf{u} - \mathbf{p}\|} \cdot \frac{\left< \mathbf{z}', \mathbf{z} \right>}{\|\mathbf{z}'\|\|\mathbf{z}\|} \label{eqn:kernel_cosin_1} 
\end{align}
For $\mathbf{z}' = \boldsymbol{\mu}_k$, we observe that $\mathbf{p}'_W$ is approximately one-hot vector with $(\mathbf{p}'_{W})_k = 1$. This observation is consistent with different architectures (DenseNet and ResNet) and training datasets (CIFAR-10, CIFAR-100, and ImageNet-1k). Substitute this one-hot vector $\mathbf{p}'$ to equation \ref{eqn:kernel_cosin_1}, we get
\begin{align*}
    K_g (\boldsymbol{\mu}_k, \mathbf{z}) &= \frac{p_k  - 1/C}{(1 - 1/C)(\|\mathbf{p}\|^2 - 1/C)} \cdot \cos(\boldsymbol{\mu}_k, \mathbf{z})
\end{align*}
Since $\mathbf{p}$ is not uniform and $p_k = {\max}_i (\mathbf{p}_W)_i$ we have $(p_k - 1/C)> 0$ and $(\|\mathbf{p}\|^2- 1/C > 0)$.
This suggests that $K_g(\boldsymbol{\mu}_k, \mathbf{z})$ and $\cos(\boldsymbol{\mu}_k, \mathbf{z})$ are positively correlated and smaller $\cos(\boldsymbol{\mu}_k, \mathbf{z})$ indicates less influence between the typical ID feature $\boldsymbol{\mu}_k$ and the test input's feature $\mathbf{z}$ and can be signal of a OOD input. 
\todo[author=hieu]{Add an toy experiment to show that the perturbation has different effect on ID vs OOD input?}

\section{Conclusion}
\label{sec:conclu}

This paper introduces \shortname, a post hoc approach to out-of-distribution detection based on the use of cosine similarity. 
The comprehensive experimental evaluation on 3 ID datasets, 10 OOD datasets, and using 3 performance metrics proves the efficacy of \shortname. 
The importance of both feature embedding and an appropriate similarity measure for effective OOD detection was also confirmed through experiments. After theoretical analysis, we have demonstrated that cosine similarity is a suitable indicator for identifying OOD samples.
We hope our work will encourage future research into using angular information for OOD detection.

\bibliography{example_paper}
\bibliographystyle{icml2023}

\newpage
\appendix
\onecolumn
\section{Implimentation}
\paragraph{Software and Hardware.} All experiments are run with PyTorch and NVIDIA RTX3090 GPUs. 

\paragraph{Number of evaluation.} In each run, we select a subset from the OOD dataset such that its size is equal to the size of the ID dataset. We run 5 times for each combination of method, ID data, and OOD data and report the average result.

Note on CIFAR results
\begin{itemize}
    \item For the reimplementation of \texttt{KNN} with DenseNet-101, we tried a grid search, where the number of neigboors $k$ include $\{10, 20, 50, 100, 200, 400\}$. We report the best result out of all hyperparameter combinations, given by $k = 50$ for CIFAR-10 and $k = 200$ for CIFAR-100. 
\end{itemize}

\section{Detailed CIFAR results}
\label{appsec:detail_cifar}
Table \ref{tab:detailresultscifar10} and \ref{tab:detailresultscifar100} present the full results on 6 OOD datasets for CIFAR-10 and CIFAR-100 benchmarks respectively.

\begin{sidewaystable}
\centering
\caption{\small Detailed results on six common OOD benchmark datasets: \texttt{Textures}~\cite{cimpoi2014describing}, \texttt{SVHN}~\cite{netzer2011reading}, \texttt{Places365}~\cite{zhou2017places}, \texttt{LSUN-Crop}~\cite{yu2015lsun}, \texttt{LSUN-Resize}~\cite{yu2015lsun}, and \texttt{iSUN}~\cite{xu2015turkergaze}. For each ID dataset, we use the same DenseNet pretrained on \textbf{CIFAR-10}. $\uparrow$ indicates larger values are better and $\downarrow$ indicates smaller values are better.}
\scalebox{0.9}{
\begin{tabular}{lllllllllllllll} \toprule
\multirow{2}{*}{\textbf{Method}} & \multicolumn{2}{c}{\textbf{SVHN}} & \multicolumn{2}{c}{\textbf{LSUN-c}} & \multicolumn{2}{c}{\textbf{LSUN-r}} & \multicolumn{2}{c}{\textbf{iSUN}} & \multicolumn{2}{c}{\textbf{Textures}} & \multicolumn{2}{c}{\textbf{Places365}} & \multicolumn{2}{c}{\textbf{Average}} \\ \cline{2-15}
 & \textbf{FPR95} & \textbf{AUROC} & \textbf{FPR95} & \textbf{AUROC} & \textbf{FPR95} & \textbf{AUROC} & \textbf{FPR95} & \textbf{AUROC} & \textbf{FPR95} & \textbf{AUROC} & \textbf{FPR95} & \textbf{AUROC} & \textbf{FPR95} & \textbf{AUROC} \\ 
 & \quad $\downarrow$ & \quad $\uparrow$ & \quad $\downarrow$ & \quad $\uparrow$ & \quad  $\downarrow$ & \quad $\uparrow$ & \quad $\downarrow$ & \quad  $\uparrow$ &  \quad $\downarrow$ & \quad $\uparrow$ & \quad $\downarrow$ & \quad $\uparrow$ & \quad $\downarrow$ & \quad $\uparrow$ \\ \midrule
Softmax score  & 47.24 & 93.48 & 33.57 & 95.54 & 42.10 & 94.51 & 42.31 & 94.52 & 64.15 & 88.15 & 63.02 & 88.57 & 48.73 & 92.46 \\
ODIN  & 25.29 & 94.57 & 4.70 & 98.86 & 3.09 & 99.02 & 3.98 & 98.90 & 57.50 & 82.38 & 52.85 & 88.55 & 24.57 & 93.71 \\
Mahalanobis  & 6.42 & 98.31 & 56.55 & 86.96 & 9.14 & 97.09 & 9.78 & 97.25 & 21.51 & 92.15 & 85.14 & 63.15 & 31.42 & 89.15 \\
Energy score & 40.61 & 93.99 & 3.81 & 99.15 & 9.28 & 98.12 & 10.07 & 98.07 & 56.12 & 86.43 & 39.40 & 91.64 & 26.55 & 94.57 \\ 
KNN & 4.14 & 99.22 & 6.97 & 98.74 & 11.26 & 97.88 & 11.50 & 98.03 & 20.48 & 96.13 & 45.28 & 90.25 & 16.61 & 96.71 \\
\shortname & 5.14 & 99.08 & 10.94 & 98.11 & 9.71 & 98.30 & 10.41 & 98.11 & 17.62 & 96.70 & 55.56 & 88.11 & 18.23 & 96.40 \\ 
\\ \bottomrule
\end{tabular}}
\label{tab:detailresultscifar10}
\end{sidewaystable}

\begin{sidewaystable}
\caption{\small Detailed results on six common OOD benchmark datasets: \texttt{Textures}~\cite{cimpoi2014describing}, \texttt{SVHN}~\cite{netzer2011reading}, \texttt{Places365}~\cite{zhou2017places}, \texttt{LSUN-Crop}~\cite{yu2015lsun}, \texttt{LSUN-Resize}~\cite{yu2015lsun}, and \texttt{iSUN}~\cite{xu2015turkergaze}. For each ID dataset, we use the same DenseNet pretrained on \textbf{CIFAR-100}. $\uparrow$ indicates larger values are better and $\downarrow$ indicates smaller values are better. }
\scalebox{0.9}{
\begin{tabular}{lllllllllllllll} \toprule
\multirow{2}{*}{\textbf{Method}} & \multicolumn{2}{c}{\textbf{SVHN}} & \multicolumn{2}{c}{\textbf{LSUN-c}} & \multicolumn{2}{c}{\textbf{LSUN-r}} & \multicolumn{2}{c}{\textbf{iSUN}} & \multicolumn{2}{c}{\textbf{Textures}} & \multicolumn{2}{c}{\textbf{Places365}} & \multicolumn{2}{c}{\textbf{Average}} \\ \cline{2-15}
 & \textbf{FPR95} & \textbf{AUROC} & \textbf{FPR95} & \textbf{AUROC} & \textbf{FPR95} & \textbf{AUROC} & \textbf{FPR95} & \textbf{AUROC} & \textbf{FPR95} & \textbf{AUROC} & \textbf{FPR95} & \textbf{AUROC} & \textbf{FPR95} & \textbf{AUROC} \\ 
 & \quad $\downarrow$ & \quad $\uparrow$ & \quad $\downarrow$ & \quad $\uparrow$ & \quad  $\downarrow$ & \quad $\uparrow$ & \quad $\downarrow$ & \quad  $\uparrow$ &  \quad $\downarrow$ & \quad $\uparrow$ & \quad $\downarrow$ & \quad $\uparrow$ & \quad $\downarrow$ & \quad $\uparrow$ \\ \midrule
Softmax score & 81.70 & 75.40 & 60.49 & 85.60 & 85.24 & 69.18 & 85.99 & 70.17 & 84.79 & 71.48 & 82.55 & 74.31 & 80.13 & 74.36 \\
ODIN & 41.35 & 92.65 & 10.54 & 97.93 & 65.22 & 84.22 & 67.05 & 83.84 & 82.34 & 71.48 & 82.32 & 76.84 & 58.14 & 84.49 \\
Mahalanobis & 22.44 & 95.67 & 68.90 & 86.30 & 23.07 & 94.20 & 31.38 & 93.21 & 62.39 & 79.39 & 92.66 & 61.39 & 55.37 & 82.73 \\
Energy score & 87.46 & 81.85 & 14.72 & 97.43 & 70.65 & 80.14 & 74.54 & 78.95 & 84.15 & 71.03 & 79.20 & 77.72 & 68.45 & 81.19 \\ 
KNN & 17.93 & 96.35 & 31.44 & 92.85 & 47.31 & 90.41 & 39.70 & 91.90 & 24.27 & 93.70 & 93.41 & 60.13 & 42.34 & 87.56 \\
\shortname & 10.03 & 97.87 & 31.90 & 93.16 & 54.41 & 88.11 & 45.43 & 90.28 & 20.43 & 95.44 & 88.38 & 69.82 & 41.76 & 89.11 \\
\\ \bottomrule
\end{tabular}}
\label{tab:detailresultscifar100}
\end{sidewaystable}

\section{Additional results}
We include the result using MobileNetV2 \cite{sandler2018mobilenetv2} for the ImageNet experiment in table \ref{tab:imagenet_mobilenetv2}.

\begin{table}[hbt!]
\caption{\textbf{OOD detection results on ImageNet.} Proposed and baseline methods are based on a MobilenetV2 model trained on ImageNet-1k \cite{deng2009imagenet} only. $\downarrow$ indicates smaller values are better and  $\uparrow$ indicates larger values are better.}
\label{tab:imagenet_mobilenetv2}
\resizebox{\textwidth}{!}
{
    \begin{tabular}{c c c c c c c c c c c}
    \hline
     & \multicolumn{8}{c}{\textbf{OOD Datasets}} & & \\
    \cline{2-9} \\
    \textbf{Methods} & \multicolumn{2}{c}{\textbf{iNaturalist}} & \multicolumn{2}{c}{\textbf{SUN}} & \multicolumn{2}{c}{\textbf{Places}} & \multicolumn{2}{c}{\textbf{Textures}} & \multicolumn{2}{c}{\textbf{Average}} \\
    & & & & & & & & & & \\
    & FPR95 & AUROC & FPR95 & AUROC & FPR95 & AUROC & FPR95 & AUROC & FPR95 & AUROC \\
    & $\downarrow$ & $\uparrow$ & $\downarrow$ & $\uparrow$ & $\downarrow$ & $\uparrow$ & $\downarrow$ & $\uparrow$ & $\downarrow$ & $\uparrow$ \\
    \midrule
    \multicolumn{1}{l}{Softmax score} & 64.29 & 85.32 & 77.02 & 77.10 & 79.23 & 76.27 & 73.51 & 77.30 & 73.51 & 79.00 \\
    \multicolumn{1}{l}{ODIN} & 55.39 & 87.62 & 54.07 & 85.88 & 57.36 & 84.71 & 49.96 & 85.03 & 54.20 & 85.81 \\
    \multicolumn{1}{l}{Mahalanobis} & 62.11 & 81.00 & 47.82 & 86.33 & 52.09 & 83.63 & 92.38 & 33.06 & 63.60 & 71.01 \\
    \multicolumn{1}{l}{Energy score} & 59.50 & 88.91 & 62.65 & 84.50 & 69.37 & 81.19 & 58.05 & 85.03 & 62.39 & 84.91 \\
    \midrule
    \multicolumn{1}{l}{\shortname\;(Our)}  & 46.66 & 90.41 & 71.28 & 77.55 & 78.86 & 73.03 & 14.49 & 96.60 & 52.82 & 84.39 \\
    \bottomrule
    \end{tabular}
}
\vspace{0.1in}

\end{table}

\section{Feature layers}
\begin{figure}[ht]
\vskip 0.2in
\begin{center}
\centerline{\includegraphics[width=0.7\columnwidth]{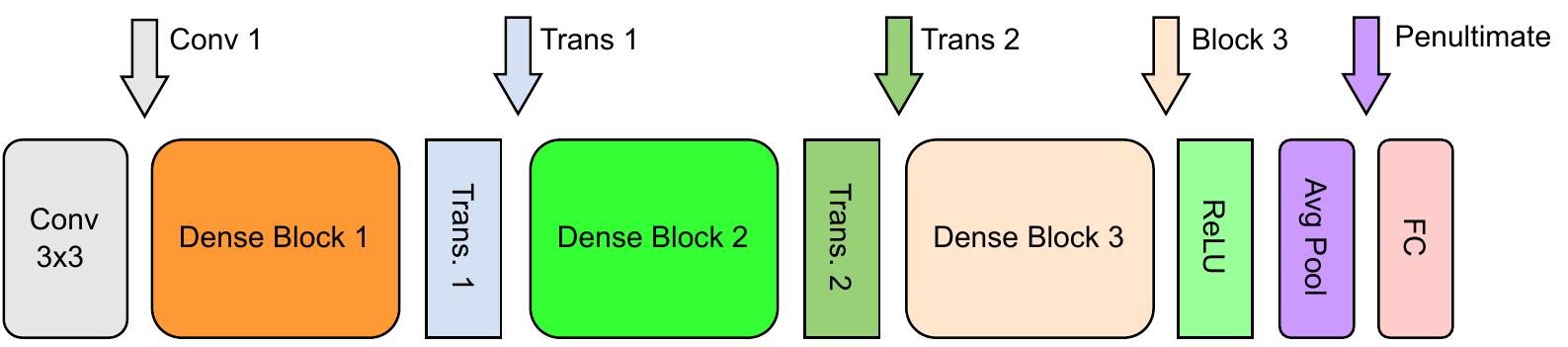}}
\caption{Diagram of DenseNet-101 architecture and indications of feature extraction layers.}
\label{fig:densenet_arch}
\end{center}
\vskip -0.2in
\end{figure}

The main results shown in the paper are generated by using features from the penultimate layer.
In table \ref{tab:cifar_layers}, we show if we use feature from different layers of the network (Illustrated in Figure \ref{fig:densenet_arch}). 

\begin{table}[hbt!]
\caption{\textbf{OOD detection results on CIFAR benchmarks.} The results were  averaged from 6 OOD datasets and measured in terms of FPR95 and AUROC. All values are percentages. All methods are based on a DenseNet-101 \cite{huang2017densely} model trained on ID data only.}
\label{tab:cifar_layers}
\centering 
\scalebox{1.0}{
\begin{tabular}{l c c | c c}
\toprule
 & \multicolumn{2}{c}{\textbf{CIFAR-10}} & \multicolumn{2}{c}{\textbf{CIFAR-100}} \\
 \cmidrule(r){2-5}
\textbf{Method} & \textbf{FPR95} {$\downarrow$} & \textbf{AUROC} $\uparrow$ & \textbf{FPR95} $\downarrow$ & \textbf{AUROC} $\uparrow$ \\
\midrule
\shortname\;before dense blocks  & 95.07 & 53.59 & 95.20 & 54.79  \\
\shortname\;after 1st Transition Block & 66.10 & 81.73 & 68.21 & 76.15 \\
\shortname\;after 2nd Transition Block & 25.91 & 92.24 & 62.87 & 78.09  \\
\shortname\;after 3rd Dense Block & 16.29 & 96.25 & 58.69 &  80.35 \\
\shortname\;after penultimate Layer & 18.23 & 96.40 & 41.76 & 89.11 \\
\bottomrule
\end{tabular}
}
\vspace{0.1in}
\end{table}

\end{document}